# A Neural Network and Iterative Optimization Hybrid for Dempster-Shafer Clustering

Johan Schubert

Department of Information System Technology
Division of Command and Control Warfare Technology
Defence Research Establishment
SE–172 90 Stockholm, Sweden
schubert@sto.foa.se

In this paper we extend an earlier result within Dempster-Shafer theory ["Fast Dempster-Shafer Clustering Using a Neural Network Structure," in *Proc. Seventh Int. Conf. Information Processing and Management of Uncertainty in Knowledge-Based Systems* (IPMU'98)] where a large number of pieces of evidence are clustered into subsets by a neural network structure. The clustering is done by minimizing a metaconflict function. Previously we developed a method based on iterative optimization. While the neural method had a much lower computation time than iterative optimization its average clustering performance was not as good. Here, we develop a hybrid of the two methods. We let the neural structure do the initial clustering in order to achieve a high computational performance. Its solution is fed as the initial state to the iterative optimization in order to improve the clustering performance.

## 1. Introduction

In this paper we develop a neural network and iterative optimization hybrid for clustering evidence in large scale problems within Dempster-Shafer theory [5]. The clustering is done by minimizing a metaconflict function. The studied problem concerns the situation when we are reasoning with multiple events which should be handled independently. We use the clustering process to separate the evidence into subsets that will be handled separately.

In earlier work [2–3] we developed a method based on iterative optimization for the clustering of evidence in medium sized problems. For large scale problems it became clear that we need a method with much lower computation time.

In a subsequent paper [4] we developed a method based on clustering with a neural structure. Here we used the structure of a neural network. There was no learning phase to set the weights. Instead, all the weights were set directly by a method where we used the conflict in Dempster's rule as input. While this method offered a great improvement in computational complexity its clustering performance was not as good.

The idea to use a neural network for optimization was inspired by a solution to the traveling salesman problem by Hopfield and Tank [1].

Here, a hybrid of the two methods is developed in order to achieve the high computational performance of the neural structure and the superior clustering performance of iterative optimization. We let the neural structure do the initial clustering. For large scale problems this is much faster than iterative optimization. However, its clustering performance is not optimal. Also, the neural structure has to work with an approximation of the metaconflict function that is being minimized. The solution found by the neural structure is fed as the initial state to the iterative optimization. Since the initial state is a good starting point, the iterative optimization finds a minimum to the metaconflict function in just a few iterations. Of course, the iterative optimization guarantees local but not global minimum.

In section 2 we describe the problem at hand. In section 3 we continue with the iterative optimization approach to solving the problem. The neural structure for clustering is described in section 4, and the neural-iterative hybrid in section 5. In section 6 we compare results of clustering performance and computational complexity of the three methods. Finally, in section 7, conclusions are drawn.

## 2. The Problem

If we receive several pieces of evidence about different and separate events and the pieces of evidence are mixed up, we want to arrange the them according to which event they are referring to. Thus, we partition the set of all pieces of evidence χ into subsets where each subset refers to a particular event. In figure 1 these subsets are denoted by $\chi_i$ and the conflict when all pieces of evidence in $\chi_i$ are combined by Dempster's rule is denoted by $c_i$. Here, thirteen pieces of evidence are partitioned into four subsets. When the number of subsets is uncertain there will also be a "domain conflict" $c_0$ which is a conflict between the current hypothesis about the number of subsets and our prior belief. The partition is then simply an allocation of all pieces of evidence to the different events. Since these events do not have anything to do with each other, we will analyze them separately.

Now, if it is uncertain to which event some pieces of evidence is referring we have a problem. It could then be impossible to know directly if two different pieces of evidence are referring to the same event. We do not know if we should put them into the same subset or not. This problem is then a problem of organization. Evidence from different events that we want to analyze are unfortunately mixed up and we are facing a problem in separating them.

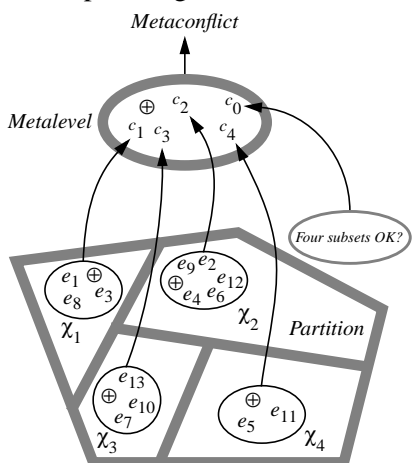

*Fig 1. The conflict in each subset of the partition becomes a piece of evidence at the metalevel.*

To solve this problem, we can use the conflict in Dempster's rule when all pieces of evidence within a subset are combined, as an indication of whether these pieces of evidence belong together. The higher this conflict is, the less credible that they belong together.

Let us create an additional piece of evidence for each subset with the proposition that this is not an "adequate partition". We have a simple frame of discernment on the metalevel $\Theta = \{\text{AdP}, \neg\text{AdP}\}$, where AdP is short for "adequate partition." Let the proposition take a value equal to the conflict of the combination within the subset,

$$m_{\chi_i}(\neg\text{AdP}) \triangleq \text{Conf}(\{e_j | e_j \in \chi_i\}).$$

These new pieces of evidence, one regarding each subset, reason about the partition of the original evidence. Just so we do not confuse them with the original evidence, let us call this evidence "metalevel evidence" and let us say that its combination and the analysis of that combination take place on the "metalevel," figure 1.

We establish [2] a criterion function of overall conflict called the metaconflict function for reasoning with multiple events. The metaconflict is derived as the plausibility that the partitioning is correct when the conflict in each subset is viewed as a piece of metalevel evidence against the partitioning of the set of evidence, χ, into the subsets, $\chi_i$.

DEFINITION. *Let the* metaconflict function,

$$Mcf(r, e_1, e_2, ..., e_n) \triangleq 1 - (1-c_0) \cdot \prod_{i=1}^{r}(1-c_i),$$

*be the conflict against a partitioning of n evidences of the set χ into r disjoint subsets $\chi_i$. Here, $c_i$ is the conflict in subset i and $c_0$ is the conflict between r subsets and propositions about possible different number of subsets.*

We will use the minimizing of the metaconflict function as the method of partitioning the evidence into subsets representing the events. This method will also handle the situation when the number of events are uncertain.

The method of finding the best partitioning is based on an iterative minimization of the metaconflict function. In each step the consequence of transferring a piece of evidence from one subset to another is investigated.

After this, each subset refers to a different event and the reasoning can take place with each event treated separately.

## 3. Iterative Optimization

For a fixed number of subsets a minimum of the metaconflict function can be found by an iterative optimization among partitionings of evidences into different subsets.

In each step of the optimization the consequence of transferring evidence from one subset to another is investigated. If a piece of evidence $e_q$ is transferred from $\chi_i$ to $\chi_j$ then the conflict in $\chi_j$, $c_j$, increases to $c_j^*$ and the conflict in $\chi_i$, $c_i$, decreases to $c_i^*$.

Given this, the metaconflict is changed to

$$Mcf^* = 1 - (1-c_0) \cdot (1-c_i^*) \cdot (1-c_j^*) \cdot \prod_{k \neq i,j}(1-c_k)$$
$$= 1 - (1-c_0) \cdot \prod_k (1-c_k).$$

The transfer of $e_q$ from $\chi_i$ to $\chi_j$ is favorable if Mcf* < Mcf. This is the case if

$$\frac{1-c_j^*}{1-c_j} > \frac{1-c_i}{1-c_i^*}.$$

It is, of course, most favorable to transfer $e_q$ to $\chi_k$, $k \neq i$, where Mcf* is minimal.

When several different pieces of evidence may be favorably transferred it will be most favorable to transfer the evidence $e_q$ that minimizes Mcf*.

It should be remembered that this analysis concerns the situation where only one piece of evidence is transferred from one subset to another. It may not be favorable at all to simultaneously transfer two or more pieces of evidence which are deemed favorable for individual transfer.

The algorithm, like all hill-climbing–like algorithms, guarantees finding a local but not a global optimum.

## 4. Neural Structure

We will study a series of problems where $2^n - 1$ pieces of evidence, all simple support functions with elements from $2^\Theta$, are clustered into $n$ clusters, where $\Theta = \{1, 2, 3, ..., n\}$.

Thus, there is always a global minimum to the metaconflict function equal to zero, since we can take all pieces of evidence that includes the 1–element and put them into cluster 1, of the remaining evidence take all those that includes the 2–element and put them into subset 2, and so forth.

The reason we choose a problem where the minimum metaconflict is zero is that it makes a good test example for evaluating performance. If another problem had been used we would have no knowledge of the global minimum and evaluation would be more difficult. We have no reason to believe that this choice of test examples is atypical with respect to network performance.

We will choose an architecture that minimizes a sum. Thus, we have to make some change to the function that we want to minimize. If we take the logarithm of one minus the metaconflict function, we can change from minimizing Mcf to minimizing a sum.

Let us change the minimization as follows

$$\min Mcf = \min 1 - \prod_i (1-c_i)$$
$$\max 1 - Mcf = \max \prod_i (1-c_i)$$
$$\max \log(1 - Mcf) = \max \log \prod_i (1-c_i)$$
$$= \max \sum_i \log(1-c_i) = \min \sum_i -\log(1-c_i)$$

where $-\log(1-c_i) \in [0, \infty]$ is a weight [5, p. 77] of evidence, i.e., metaconflict.

Since the minimum of Mcf (= 0) is obtained when the final sum is minimal (= 0) the minimization of the final sum yields the same result as a minimization of Mcf would have.

Thus, in the neural network we will not let the weights be directly dependent on the conflicts between different pieces of evidence but rather on $-\log(1 - c_{jk})$, where $c_{jk}$ is the conflict between the $j$th and $k$th piece of evidence;

$$c_{jk} = \begin{cases} m_j \cdot m_k, & \text{conflict} \\ 0, & \text{no conflict.} \end{cases}$$

This, however, is a slight simplification since the neural structure will now minimize a sum of $-\log(1 - c_{jk})$, but take no account of higher order terms in the conflict.

Let us study the calculations taking place in the neural network during an iteration. We use the same terminology as Hopfield and Tank [1] with input voltages as the weighted sum of input signals to a neuron, output voltages as the output signal from a neuron, and inhibition terms as negative weights.

For each neuron $n_{mn}$ we calculate an input voltage u as the weighted sum of all signals from row *m* and column *n*, figure 2.

This sum is the previous input voltage of the previous iteration for $n_{mn}$ plus a gain factor times the sum of the weighted sum of output voltages $V_{ij}$ of all neurons of the same column or row as $n_{mn}$ plus an excitation bias and minus the previous input voltage of $n_{mn}$.

Thus, the new input voltage to $n_{mn}$ at iteration $t + 1$ is

$$u_{mn}^{t+1} = u_{mn}^{t} + \eta \cdot \left( \sum_{i} [-\text{dti} \cdot \log(1 - c_{in}) + \text{gi}] \cdot V_{in} + \sum_{j \neq n} [\text{ri} + \text{gi}] \cdot V_{mj} + \text{eb} - u_{mn}^{t} \right)$$

where $\eta$ is the gain factor.

From the new input voltage to $n_{mn}$ we can calculate a new output voltage of $n_{mn}$

$$V_{mn}^{t+1} = \frac{1}{2} \cdot \left( 1 + \tanh(\frac{u_{mn}^{t+1}}{u_0}) \right)$$

where tanh is the hyperbolic tangent, $u_0 = 0.02$, and $V_{mn}^{t+1} \in [0,1]$.

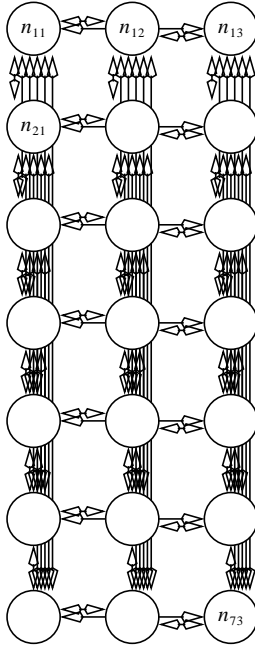

*Fig 2. Neural network. Each column corresponds to a cluster and each row corresponds to a piece of evidence.*

Initially, before the iteration begins, each neuron is initiated with an input voltage of $u_{00}$ + noise where

$$u_{00} = u_0 \cdot \text{atanh}\left(\frac{2}{n} - 1\right)$$

and atanh is the hyperbolic arc tangent.

The initial input voltage is set at $u_{00} + \delta u$ where $\delta u$, the noise, is a random number chosen uniformly in the interval $-0.1 \cdot u_0 \leq \delta u \leq 0.1 \cdot u_0$.

After convergence is achieved the conflict within each cluster, i.e., column, is calculated by combining those pieces of evidence for which the output voltage for the column is 1.0.

We now have a conflict for each subset and can calculate the overall metaconflict, Mcf, by the previous formula.

## 5. Neural-iterative Hybrid

The idea behind the neural-iterative hybrid is simple. We want to develop a method that has the superior computational complexity of the neural clustering and at the same time has the excellent clustering performance of iterative optimization. This means that most of the actual work has to be done by the neural part of the hybrid in order to achieve a low computational time. Only when we have almost found a solution can we afford to use iterative optimization to achieve maximum clustering performance.

In this hybrid the neural structure does the first clustering until it reaches convergence. The solution found by the neural structure is then fed as the initial state to the iterative optimization. Iterative optimization takes place and continues until convergence is achieved.

In figure 3 a typical convergence of a neural network and the following iterative optimization of 63 pieces of evidence into six clusters is seen. It takes the neural structure 71 iterations to converge. This is viewed at its first, eleventh, 21st, 31st, ..., 61st and 71st iteration. The final state of the neural clustering is taken as the initial state of the iterative optimization. Here, in each iteration one piece of evidence is moved from one cluster to another. The iterative optimizations takes three more iterations (the 72nd – 74th iteration). In figure 3 they are market grey with the changes marked black.

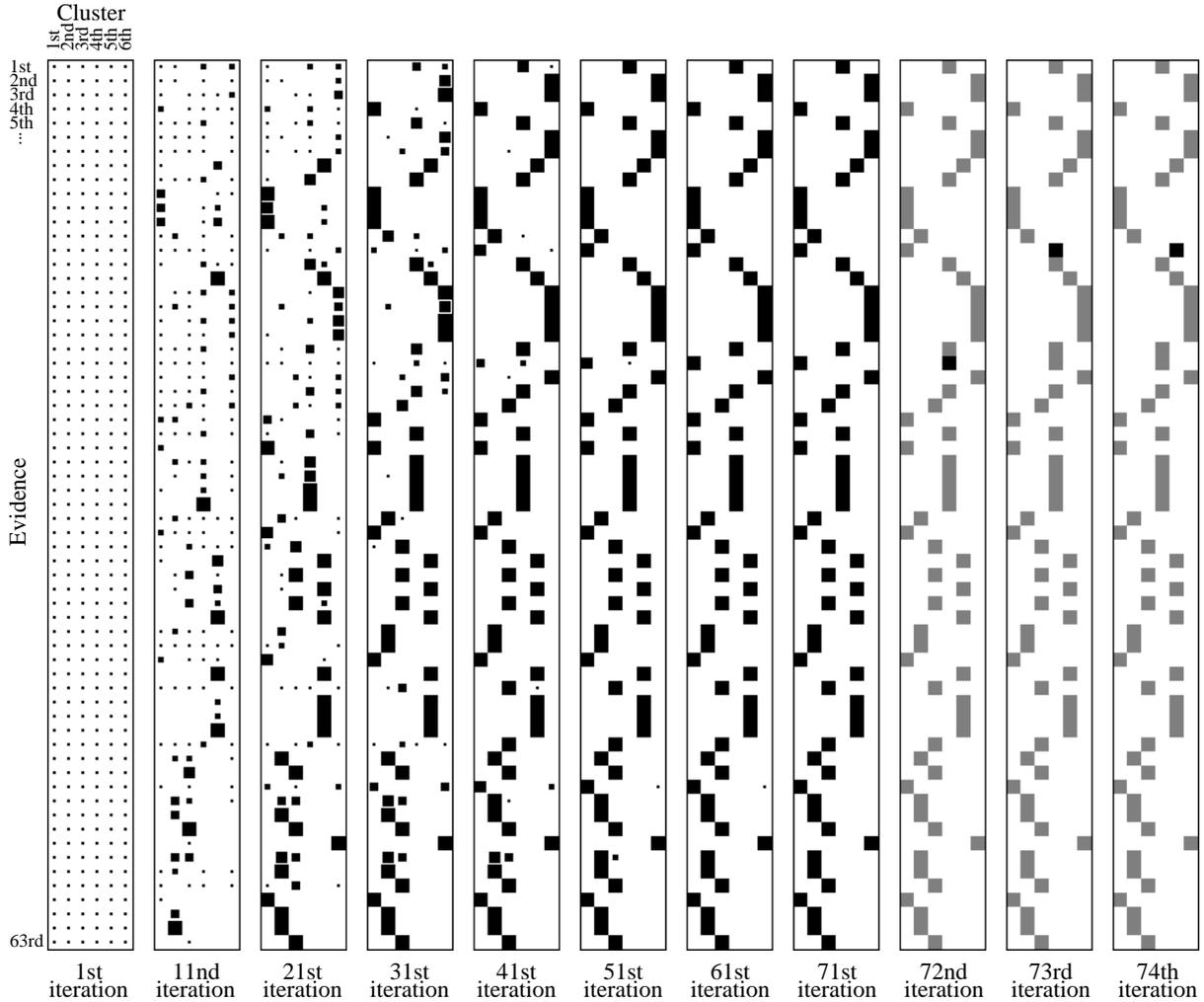

*Fig 3. Neural-iterative hybrid: The convergence of a neural network (iterations 1-71, black) followed by iterative optimization (iterations 72-74, grey). In each snap-shot of an iteration each column represents a cluster and each row represents a piece of evidence. The linear dimension of each square is proportional to the output voltage of the neuron and represent the degree to which a piece of evidence belong to a cluster. During iterative optimization one piece of evidence is moved in each iteration from one cluster to another (black) until convergence.*

The clustering performance can be seen in figure 4. Here, the metaconflict from all clusters are calculated for each iteration. It should be noticed that the neural structure part of the hybrid method minimizes an approximation of the metaconflict function, while the iterative part minimizes the actual function. This explains the uphill curve between the 37th and 71st iteration until convergence is reached.

One observation is that most of the neural structures clustering performance is achieved after half of all iterations. Thus, it might be possible to take an intermediate solution of the neural structure and feed that one to the iterative optimization, although that has not been done here. However, repeated calculation of metaconflict at each iteration is time consuming and not a good idea, and taking a "half ready" solution of the neural structure might be counter productive because of the high computational complexity of the following iterative optimization part. In the investigation of computational complexity and cluster performance in the next section we always wait for convergence of both parts of the hybrid method.

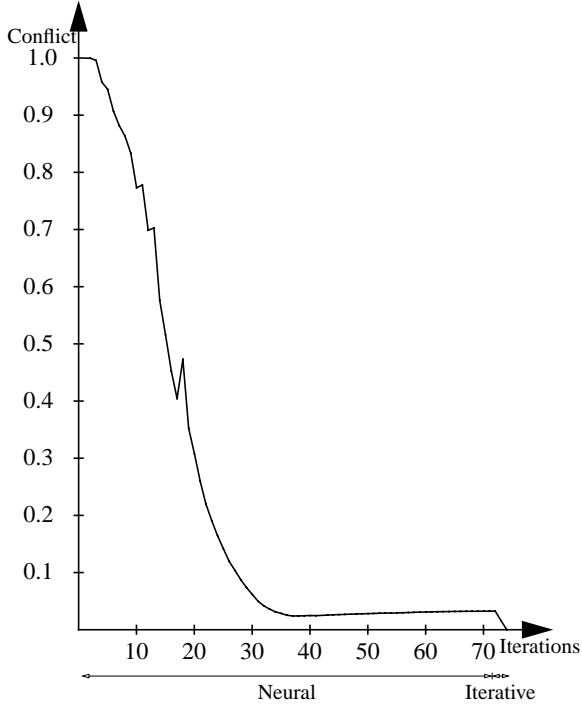

*Fig 4. Metaconflict per iteration for the "63 pieces into 6 clusters" problem of the neural-iterative hybrid.*

## 6. Results

### 6.1   $2^n - 1$ into $n$ problems

Let us investigate clustering performance and computational complexity of the three clustering processes; the neural structure, iterative optimization and the neural-iterative hybrid. We make this comparison as the problem size grows.

For all problem sizes we will cluster $2^n - 1$ pieces of evidence into $n$ different clusters. In this test we use evidence that support the different subsets of the frame $\Theta = \{1, 2, 3, ..., n\}$. Thus, we know that the metaconflict function has a global minimum with metaconflict equal to zero.

In table 1 and figure 5 we see the exponential growth in computation time of the iterative optimization, the neural structure and the hybrid method as the problem size grows. The neural structure has a much lower computational complexity than iterative optimization, although it is still exponential, but it has a higher computation time for small problems.

For the neural-iterative hybrid we notice that its computational complexity is very close to that of neural clustering. For instance, in the "63 pieces into 6 clusters" problem the computation time is 120 seconds for the hybrid method compared to 109 seconds of the neural, table 1. Note also, that on average there are only 3.5 extra iterations in the iterative optimization part of the hybrid clustering after the initial neural clustering has converged. Compared with the 26.1 iterations of the iterative optimization this indicates that the neural structure of the hybrid method does most of the job, but to achieve high performance in clustering we need the iterative optimization to finish the job. The same conclusion can be drawn from figure 4.

We should also notice that the ∆-time difference between hybrid and neural methods grows. For very large scale problems where we can anticipate a moderately successful clustering performance of the neural method, its solution will not be a sufficiently good starting state for the iterative part of the hybrid method. Thus, we expect the computational complexity for very large scale problems to approach that of iterative

Table 1: Computation time and iterations

| # Evidence | 7 | 15 | 31 | 63 | 127 |
|---|---|---|---|---|---|
| # Clusters | 3 | 4 | 5 | 6 | 7 |
| Neural structure | | | | | |
| time (s) | 2.09 | 7.39 | 27.3 | 109 | 618 |
| iterations | 54.3 | 63.7 | 65.2 | 79.8 | 108 |
| Iterative optimization | | | | | |
| time (s) | 0.061 | 0.201 | 1.90 | 288 | 76d* |
| iterations | 2.6 | 5.1 | 11.1 | 26.1 | – |
| Hybrid | | | | | |
| time (s) | 2.31 | 8.35 | 28.2 | 120 | – |
| ∆-time (s) | 0.22 | 0.96 | 0.91 | 11.4 | – |
| ∆ iterations | 1.5 | 2.4 | 3.4 | 3.5 | – |

*estimated (days)

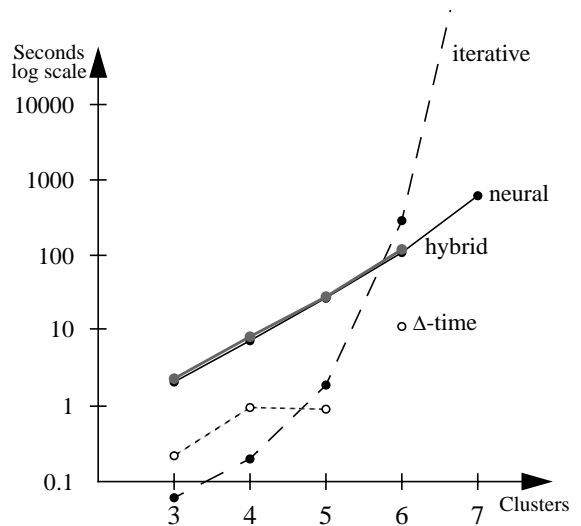

*Fig 5. Computation time (mean) of neural, iterative, hybrid and ∆-time: hybrid-neural.*

optimization. We found in [4] that for these very large scale problems some heuristic preclustering might be necessary. The same conclusion applies here.

Having found the computational complexity of the hybrid method to be almost as good as that of the neural structure we will now study the clustering performance of the three methods. In table 2 and figure 6 we find the metaconflict of the three methods as the problem size grows from three clusters to six. For each problem size $n$ we have clustered the $2^n - 1$ different pieces of evidence from the set of all subsets of Θ ten times with different random basic probability numbers each time.

While all three methods manage to find a global minimum at least one time out of ten for all problem sizes, we see that the mean conflict over ten runs is vastly better for the hybrid and iterative methods compared to neural clustering. The hybrid method is slightly better than iterative optimization.

In [4] we said that cluster performance should be measured not by metaconflict in it self since the difficulty of the problem grows with the problem size. Instead, we prefer to measure the performance by conflict per cluster or conflict per cluster and evidence, the latter being the best measure of clustering performance.

In table 3 and figures 7 and 8 we see a good performance on clustering by the hybrid method. It is slightly better than that of iterative optimization and much better than neural clustering.

While the clustering performance of the hybrid method is found to be better than that of iterative optimization and the computational complexity is almost as good as that of neural clustering, we have not yet found the problem size limit of the method. This is done below by studying the six-cluster problem for successively larger problem sizes.

### 6.2 The six-cluster problem

Let us further study the six-cluster problem. Here we draw random subsets from the set of all subsets as propositions with random basic probability numbers attached.

In table 4 and figure 9 we see a good performance of the hybrid model up to 70

Table 2: Metaconflict

| # Evidence | 7 | 15 | 31 | 63 |
|---|---|---|---|---|
| # Clusters | 3 | 4 | 5 | 6 |
| Neural structure | | | | |
| best of 10 | 0 | 0 | 0 | 0 |
| mean | 0.016 | 0.059 | 0.076 | 0.398 |
| Iterative optimization | | | | |
| best of 10 | 0 | 0 | 0 | 0 |
| mean | 0 | 0.001 | 0.003 | 0.097 |
| Hybrid | | | | |
| best of 10 | 0 | 0 | 0 | 0 |
| mean | 0 | 0 | 0.002 | 0.090 |

Table 3: Conflict per cluster and evidence

| # Evidence | 7 | 15 | 31 | 63 |
|---|---|---|---|---|
| # Clusters | 3 | 4 | 5 | 6 |
| Neural structure (mean conflict) | | | | |
| / cluster | 0.005 | 0.015 | 0.016 | 0.081 |
| / evidence | 0.002 | 0.004 | 0.003 | 0.008 |
| Iterative (mean conflict) | | | | |
| / cluster | 0 | 0.0003 | 0.0005 | 0.017 |
| / evidence | 0 | 0.00009 | 0.00008 | 0.002 |
| Hybrid (mean conflict) | | | | |
| / cluster | 0 | 0 | 0.0004 | 0.016 |
| / evidence | 0 | 0 | 0.00007 | 0.001 |

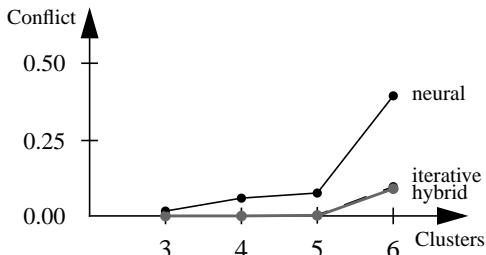

Fig 6. *Metaconflict (mean) of neural, iterative and hybrid methods.*

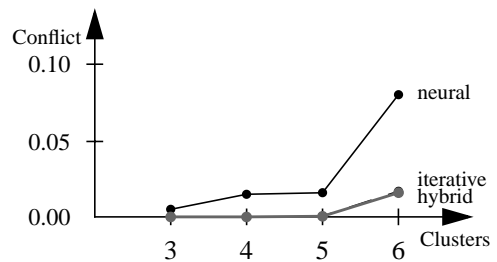

Fig 7. *Metaconflict per cluster.*

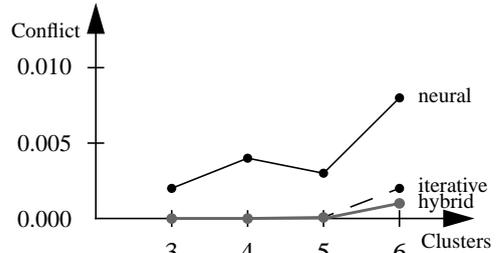

Fig 8. *Metaconflict per cluster and evidence.*

Table 4: Computation time and iterations

| # Evidence | 50 | 60 | 70 |
|---|---|---|---|
| Neural structure | | | |
| time (s) | 75.7 | 98.3 | 142.8 |
| iterations | 78.8 | 73.5 | 80.1 |
| Hybrid | | | |
| time (s) | 80.8 | 117.4 | 268.2 |
| Δ-time (s) | 5.1 | 19.1 | 125.4 |
| Δ iterations | 5.2 | 5.8 | 7.4 |

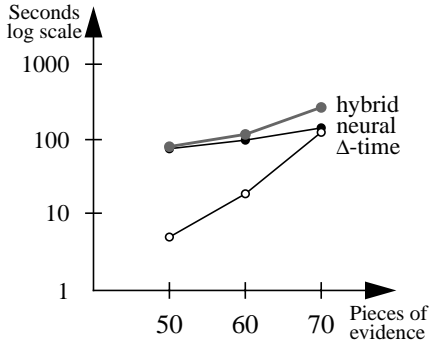

Fig 9. *Computation time of the hybrid and the neural method in the six-cluster problem.*

pieces of evidence. While the conflict of the neural part improves from 0.574 to 0.520 in the 70-pieces problem, table 5, the effort of optimizing 70-pieces of evidence lead to an almost doubling in computation time. From the increase in Δ-time we conclude that the limitation of the hybrid method lies somewhat beyond 70-pieces.

In table 5 and figure 10 we notice the vastly superior result of the hybrid method compared to the neural optimization. Notice also, that the hybrid method found the global minimum for each problem size, in the 50-pieces problem every time, in the 60-pieces problem seven out of ten times, and in the 70-pieces problem four out of ten times (not shown in table).

Finally, in table 6, we see the conflict per cluster and conflict per piece of evidence compared for the two methods. For instance, in the 70-pieces problem, we have a conflict per piece of evidence of 1/50th for the hybrid method compared to the neural optimization.

Table 5: Conflict

| # Evidence | 50 | 60 | 70 |
|---|---|---|---|
| Neural structure | | | |
| best of 10 | 0.254 | 0.050 | 0.052 |
| mean | 0.494 | 0.574 | 0.520 |
| Hybrid | | | |
| best of 10 | 0 | 0 | 0 |
| mean | 0 | 0.005 | 0.015 |

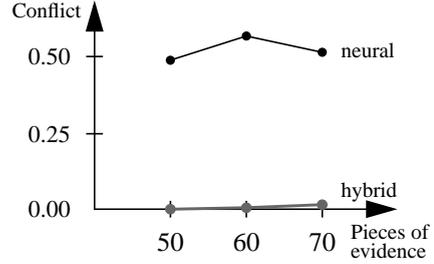

Fig 10. *Conflict of hybrid and neural methods.*

Table 6: Conflict per cluster and evidence

| # Evidence | 50 | 60 | 70 |
|---|---|---|---|
| Neural structure (mean) | | | |
| / cluster | 0.107 | 0.133 | 0.092 |
| / evidence | 0.013 | 0.013 | 0.010 |
| Hybrid (mean) | | | |
| / cluster | 0 | 0.0009 | 0.003 |
| / evidence | 0 | 0.00009 | 0.0002 |

## 7. Conclusions

We have shown the new hybrid method to have practically the computational complexity of neural clustering and the clustering performance of iterative optimization. This holds true for medium to large scale problems. For very large problems the clustering performance of the initial neural part of the hybrid method is not sufficiently good as an initial state for the subsequent iterative optimization part to yield a low computation time.